%% file: main.tex
\documentclass[sigconf]{acmart}
\usepackage{booktabs}
\usepackage{amsmath}
\usepackage{microtype}
\usepackage{tikz}
\usetikzlibrary{arrows.meta,positioning}
\setcopyright{none}
\renewcommand\footnotetextcopyrightpermission[1]{}
\acmConference[WWW '27]{Submission to The ACM Web Conference 2027}{May 2027}{Dublin, Ireland}
\acmYear{2027}
\copyrightyear{2027}
\acmDOI{}
\acmISBN{}
\newcommand{\MarketMonths}{65,426}
\newcommand{\UniqueMarkets}{47,075}
\newcommand{\PooledEdges}{5,667}
\newcommand{\UniquePairs}{5,315}
\newcommand{\GraphNodes}{5,361}
\newcommand{\TradeRows}{30,311,603}
\newcommand{\TradeMarketMonths}{59,361}

\makeatletter
\@ACM@balancefalse
\makeatother
\newcommand{\significancenote}{$^{*}p<0.10$, $^{**}p<0.05$, $^{***}p<0.01$.}

\title[]{Agentic AI for Clustering and Relationship Discovery in Prediction Markets}
\author{Agostino Capponi}
\affiliation{\institution{Columbia University}\country{United States}}
\email{ac3827@columbia.edu}
\author{Alfio Massimiliano Gliozzo}
\affiliation{\institution{IBM Research}\country{United States}}
\email{gliozzo@us.ibm.com}
\author{Brian Zi Qi Zhu}
\affiliation{\institution{Columbia University}\country{United States}}
\email{bzz2101@columbia.edu}
\renewcommand{\shortauthors}{Capponi et al.}

\begin{abstract}
Prediction market platforms list thousands of contracts whose logical and economic connections are implicit in their resolution rules. We introduce a hierarchical agentic reduction (HAR) pipeline for constructing inspectable cross-event relation graphs and a behavioral benchmark for evaluating their alignment with market prices. This partitions a catalog of prediction markets, specified by question and resolution text, into topic clusters, applies typed LLM reasoning to small batches, and proposes signed relations with rationales between specific markets. We evaluate semantic relation graphs spanning eight monthly Kalshi catalogs (February to September 2026). Using data from 30.3 million trades, we compute a benchmark that compares linked pairs' signed price-change correlations with those of matched, unlinked controls from the same topic cluster on identical timestamps. Under full-sample and leave-first-week-out evaluation, relations identified by our pipeline exhibit higher signed Pearson price-change correlations, with statistical significance, than matched within-topic controls. This increase also appears under Spearman correlation and becomes larger among relations labeled as medium or high strength. This benchmark provides a criterion for comparing relation-discovery methods against observations beyond their input text, and our results indicate the ability of agentic AI and LLM reasoning to identify meaningful relations between individual markets in connected Web marketplaces.

\end{abstract}


\begin{document}
\maketitle

\section{Introduction}
Prediction-market contracts turn descriptions of future events into individually addressable, tradable propositions. Each contract has a question, an identifier, and a rule that specifies when its Yes outcome holds. Across a platform, these descriptions form a fragmented representation of connected uncertainties. For example, a central-bank decision and a Treasury-yield threshold may share policy exposure, resignation and early departure from office are naturally connected, and two election questions may encode competing outcomes. These relationships often cross the exchange's event boundaries as well.

Recovering this structure is a Web knowledge-integration problem with practical consequences. Explicit cross-market relations let traders and platforms detect incoherent prices across logically linked contracts, construct hedges across separately listed events, navigate and recommend related markets, and study how information propagates through a market ecosystem. Yet platforms expose contracts as isolated listings, and connections among their meanings remain implicit in free-text resolution rules. A useful graph must link propositions with different wording while respecting the jurisdictions, deadlines, and thresholds that fix their meaning, and it should be testable against evidence other than the text from which its edges were inferred. Price changes provide an external measure of a relation's behavioral relevance. Their association reflects how traders update valuations as information arrives, alongside differences in contract horizons and trading activity. Semantic validity concerns the relationship between contract propositions; the price benchmark measures how that relationship appears in observed valuation changes.



We introduce HAR for constructing inspectable cross-event relation graphs and a behavioral benchmark for evaluating their alignment with market prices. HAR groups contracts through event-title embeddings, reduces batches of markets to structured relation proposals, and assembles validated proposals into a graph.
Each edge specifies two contract identities, an expected association sign, an ordinal strength, and a rationale. The hierarchy consists of topic partitioning, batch-level reasoning, and graph assembly. 
This design makes each proposed relation inspectable and allows the model to return an empty relation set when no connection is sufficiently supported.

Our empirical question is whether selected edges, corresponding to relations between prediction markets, align with price co-movement more closely than comparable pairs drawn from the same topic cluster. 

We evaluate semantic relation graphs spanning eight monthly Kalshi catalogs (Section~\ref{sec:market-sample}). Our benchmark compares each linked pair's signed price-change correlation with that of activity-matched, unlinked controls from the same topic cluster, computed on identical timestamps. Controls are matched on the first week of each month; LFWO measures correlations over the remaining days. The evaluated graph yields signed correlation advantages of $+0.071$ for three-hour changes and $+0.033$ for daily changes under LFWO, with a larger three-hour estimate of $+0.107$ for medium/high relations. The LLM used for relation generation has a documented January 2025 knowledge cutoff~\cite{google2026cutoff}, preceding February 2025 and the 2026 evaluation period. Its inputs contain contract questions and resolution rules, excluding trade prices and settlement fields. These results describe exploratory evidence from a retrospective specification search; Section~\ref{sec:uncertainty} gives the inference scope.


We make three contributions. First, we represent cross-event propositions in a signed, attributed graph with exact contract identities, resolution conditions, and temporal scope. Second, we introduce HAR, which combines topic partitioning and structured LLM reasoning to propose individually justified relations from contract descriptions. Third, we develop a behavioral benchmark that measures signed price-correlation advantages over random and activity-matched within-topic controls using identical observations. Analyses across temporal resolutions, relation strengths, and sign assignments connect graph attributes to observed price behavior. The construction workflow and benchmark together provide a reproducible task definition for comparing relation-discovery methods.

\section{Background and Related Work}

\paragraph{Prediction markets and temporal evidence.}
Prediction markets aggregate information through contingent claims~\cite{wolfers2004prediction}. A binary market offers Yes and No contracts: a Yes contract pays \$1 when its proposition resolves affirmatively, and a No contract pays \$1 when it resolves negatively. These claims trade through mechanisms such as limit order books and automated market makers. The resolution rule ties each contract to a precisely specified proposition.

An individual contract in our catalog represents a binary proposition, while an exchange event can contain several contracts describing alternative outcomes. For example, an election event may offer separate Yes/No contracts for several candidates. When the alternatives are mutually exclusive and collectively exhaustive, their settlement indicators sum to one, creating a direct logical constraint among contracts~\cite{hanson2003combinatorial}. Other event structures permit overlapping outcomes, so their resolution rules determine the applicable constraints. Our graph keeps contract identity separate from event membership and focuses on relationships across distinct exchange events.

We study the relationships implicit across separately listed contracts. Recent LLM forecasting work evaluates probabilities against realized outcomes~\cite{halawi2024forecasting}, while ForecastBench emphasizes questions beyond the forecaster's available information~\cite{karger2025forecastbench}. 
Our correlation benchmark evaluates an association between two evolving beliefs, as a single pair of realized settlements offers limited evidence about such an association, especially when an edge expresses shared exposure rather than a strict logical constraint.

Related market-structure work combines topical embeddings, temporal filters, and LLM judgments to identify combinatorial dependencies on Polymarket~\cite{saguillo2025}. Cross-platform semantic alignment also relies on resolution conditions and temporal scope to establish event identity~\cite{gebele2026}. Consistency checks for LLM forecasters evaluate logical constraints among their predicted probabilities~\cite{paleka2025consistency}. Our benchmark asks how a signed cross-event graph, including softer associations, aligns with observed price changes under shared observation support and within-topic controls, and complements these approaches through a behavioral criterion that can be applied to alternative graph generators.

\paragraph{Knowledge graphs.}
Constructing a knowledge graph requires choices about identity, schema, and context~\cite{hogan2021knowledge}. 
These choices are especially consequential for prediction markets: two questions about the same person can be over different jurisdictions, have different deadlines or thresholds, and use different resolution sources. 
The unit of identity in our graph is the exchange contract. Event membership and topic clusters are contextual attributes, while an inferred relation is an additional, uncertain assertion about two contract propositions.

Generative relation extraction methods, such as REBEL~\cite{huguetcabot2021rebel}, map text to structured ontologies. The Extract--Define--Canonicalize method separates extraction, relation definition, and schema canonicalization for LLM-based graph construction~\cite{zhang2024edc}. 
Our setting adds an external behavioral benchmark, as the endpoints denote contracts with observable price histories.
We note that our graph structure uses signed associations and rationales, and is less semantically granular than a fully logical ontology.

\paragraph{Structured agentic workflows.}
Agentics expresses reasoning as typed transductions between structured data objects~\cite{gliozzo2025transduction}. We use its batch-reduction abstraction to transform a set of market descriptions into a selectively emitted set of edges. Compared with a question-by-question forecasting system, the semantic output is a relationship among propositions. Batching makes context size manageable and creates a candidate-coverage trade-off that belongs in the method specification.

\section{Constructing the Semantic Relation Graph}
\label{sec:construction}
\subsection{Market sample and scope}
\label{sec:market-sample}
We construct eight monthly Kalshi catalogs spanning February--September 2026. The initial market selection uses Kalshi market reports on Dune. For each report window, the query sums daily volume by ticker and category, selects ticker--category groups with cumulative reported volume of at least 10,000, and retrieves the active and finalized subsets separately. Each query returns up to 10,000 ticker--category groups ordered by volume. We combine these subsets and enrich the selected events with contract questions and resolution rules. Enrichment also includes sibling contracts within those events, so the volume threshold applies to the initial selection. Appendix~\ref{app:sample} gives the exact query conditions.

The query excludes the source categories Crypto, Sports, Financials, Exotics, and Commodities, along with the \texttt{KXHIGH}, \texttt{KXLOWT}, and \texttt{KXRAIN} ticker families for local weather outcomes. Our scope emphasizes relationships that require interpreting context and resolution language across distinct events. Repeated sports fixtures, cryptocurrency up/down questions, asset-price thresholds, and local weather ladders can produce many closely templated contracts whose connections follow directly from a common fixture, instrument, or measurement. The exclusions reduce their influence on the catalog and focus the benchmark on contextual relationships such as political developments, institutional decisions, and economic releases. The volume floor prioritizes observed market interest for the later price-based evaluation. These choices define the scope of the benchmark; the selected topics include politics, elections, economics, companies, entertainment, and science and technology.

The evaluated catalogs contain \MarketMonths{} distinct market-month observations and \UniqueMarkets{} unique tickers. 
The graph contains \PooledEdges{} monthly edges, corresponding to \UniquePairs{} distinct unordered ticker pairs and \GraphNodes{} endpoint tickers. Of the monthly edges, 4,901 are positive and 766 negative. This sign imbalance motivates the all-positive sensitivity in Appendix~\ref{app:sign}. Table~\ref{tab:coverage} reports monthly catalog and graph sizes alongside the trade-data coverage used for evaluation.

\subsection{Contract and edge representation}
For a monthly catalog $m$, we represent contract $i$ as
\begin{equation*}
 x_i^m=(t_i,e_i,q_i,r_i,\tau_i),
\end{equation*}
where $t_i$ is the ticker, $e_i$ the exchange event, $q_i$ the question, $r_i$ the resolution rule, and $\tau_i$ the available temporal metadata. The latent binary settlement proposition is $Y_i$. The LLM receives the market information: a question about an event and its resolution criteria. Note that the price series and settlement outcomes are excluded from these input fields.

The graph $G_m=(V_m,E_m)$ has ticker-addressed vertices and edge assertions
\begin{equation*}
 a_{ij}^m=(t_i,t_j,o_{ij},s_{ij},h_{ij},z_{ij}),
\end{equation*}
The proposed endpoint ordering is $o_{ij}$, and the association sign is $s_{ij}\in\{-1,+1\}$. The ordinal strength $h_{ij}$ takes the values low, medium, or high; $z_{ij}$ is the rationale.

The rationale describes the asserted mechanism under which the markets are related. An implication, a shared macroeconomic exposure, and a competitive relationship have distinct interpretations even if two of them share a sign. Strength also remains separate from relation type: a high-strength label is an ordinal judgment rather than an estimated probability of correctness. 

\begin{figure*}[t]
\centering
\begin{tikzpicture}[font=\small,>=Latex,box/.style={draw,rounded corners=2pt,align=center,text width=2.6cm,minimum height=0.95cm,fill=blue!4},line/.style={->,thick}]
\node[box] (text) at (0,1.1) {Contract questions\\and resolution rules};
\node[box] (cluster) at (3.15,1.1) {Event-title embeddings\\and topic clusters};
\node[box] (reduce) at (6.3,1.1) {Batches of up to ten\\typed LLM reduction};
\node[box] (valid) at (9.45,1.1) {Exact identity checks\\and edge assembly};
\node[box,fill=orange!7] (pool) at (12.6,1.1) {Pooled semantic\\relation graph};
\draw[line] (text)--(cluster);\draw[line] (cluster)--(reduce);\draw[line] (reduce)--(valid);\draw[line] (valid)--(pool);
\node[box,fill=red!8] (trades) at (0,-.45) {Observed trades\\February--September 2026};
\node[box,fill=red!8] (controls) at (3.15,-.45) {First-week controls\\and common timestamps};
\node[box,fill=green!5] (benchmark) at (6.3,-.45) {Sign-aligned price-change\\correlation benchmark};
\draw[line] (trades)--(controls);\draw[line] (controls)--(benchmark);
\draw[line] (pool.south) to[out=-90,in=0] (benchmark.east);
\end{tikzpicture}
\caption{Graph construction and evaluation. The top path shows the HAR contribution to the semantic relation graph. Traded-price data enters through another path, supplying controls and observations for correlation evaluation.}
\Description{Questions and resolution rules pass through topic clustering, ten-contract LLM batches, and identity validation to contribute edges to the graphs.}
\label{fig:pipeline}
\end{figure*}

\subsection{Topic clustering and relationship discovery}

\paragraph{Clustering a monthly catalog by topic.}
We embed normalized event titles using the Sentence-BERT model \texttt{all-MiniLM-L6-v2}~\cite{reimers2019sbert,minilm2021}. Unit-normalized vectors are reduced to ten dimensions by UMAP using cosine distance with thirty neighbors and a zero minimum distance~\cite{mcinnes2018umap}. We then use HDBSCAN to cluster the embeddings, requiring a minimum cluster size of four and a minimum sample size of two per cluster~\cite{mcinnes2017hdbscan}. 
Distance-weighted nearest-neighbor assignment places remaining markets, initially categorized as noise, into established clusters where possible. 
Contracts inherit their event's cluster, and cluster labels are local to a monthly catalog, with no shared cluster identity across months.

\paragraph{Within-cluster relationship discovery.}
Let $C_{mc}$ contain the contracts assigned to cluster $c$. HAR partitions its contracts into batches $B_{mck}$ of at most ten and applies the typed transducer $T_\theta$:
\begin{align*}
 \mathcal R_{mck}&=T_\theta(B_{mck};\mathcal I,\mathcal S),\\
 E_m^{H}&=\mathcal V\!\left(\bigcup_{c,k}\mathcal R_{mck}\right).
\end{align*}
Here $\mathcal I$ gives the relation-selection instructions, $\mathcal S$ declares the output fields, and $\mathcal V$ resolves identities and enforces graph constraints. A batch may emit zero edges. The model considers several propositions together and chooses which relationships to express; the output size is determined by selection rather than a required edge quota.

We require the two endpoints of a proposed relation to belong to different exchange events. Contracts within one event often share a resolution source and describe mutually exclusive outcomes or nested numerical thresholds. Their relationships can follow directly from that event's structure. The cross-event requirement focuses discovery on connections across separately listed questions, such as a policy decision and an economic release. The prompt requests different event titles, and the validator checks distinct exchange event tickers after resolving the endpoint descriptions. Topic-cluster membership supplies candidate context while this event check defines the pairs included in the evaluation.

Agentics' \texttt{areduce} implementation wraps each batch as a typed input and performs asynchronous batch transductions. Graph assembly unions their outputs. The hierarchy consists of the catalog, topic clusters, batches, and edge assertions. HAR evaluates relationships within each batch, so candidate coverage depends on the cluster partition and batch ordering. Exhaustive pairwise inference can assess pairs across batch boundaries within the same contract universe.

\subsection{Implementation details}

Our prompt for the agentic pipeline requires exact reproduction of each endpoint's question and primary resolution rule. The validator resolves these descriptions to unique exchange tickers and checks distinct market tickers, distinct event tickers, and common membership in a monthly topic cluster. For contemporaneous price correlations, each unordered market pair contributes once, regardless of the order of its two tickers. Duplicate edges are collapsed, and pairs with conflicting signs are excluded.

We use Gemini 3.1 Pro with zero temperature and a fixed seed. Google reports a January 2025 knowledge cutoff~\cite{google2026cutoff}, which precedes February 2025 and is more than a year before the evaluation period. Model inputs contain contract questions and resolution rules, excluding trade prices and settlement fields. The model reasons without Web search or news retrieval. The cutoff describes its learned information; graph construction and evaluation are retrospective, and temporal validity also depends on the versions and dates of input metadata.

\subsection{Worked example}
As an illustrative example, consider four contracts from a March 2026 cluster of questions about Donald Trump's term in office.
Their questions concern impeachment, impeachment and removal, resignation, and early departure from office. Topic similarity places them in a common search context, and resolution rules supply the distinctions needed for relation selection.
The removal contract requires impeachment \emph{and} Senate conviction before 20 January 2029, whereas the departure contract requires leaving office before that date; the resignation contract concerns the elected 2025--2029 term. 
Appendix~\ref{app:example} contains the full input descriptions and a relation example.

For removal and departure, the output assigns \texttt{direction=true} and \texttt{strength=high}, and explains that conviction and removal entail leaving office within the specified horizon. A second high-strength positive edge connects resignation to departure. These edges express distinct mechanisms linking to the same departure contract. Their rationales make the semantic basis inspectable before any correlation is calculated.

The impeachment-only contract adds a useful distinction. Its edge to removal has medium strength: House impeachment is a prerequisite for Senate conviction, and subsequent conviction remains uncertain. This rationale motivates a positive association while preserving the extra condition in the removal rule. The strength label indicates the generator's judgment about that relationship. Each edge includes its rationale alongside the label, allowing readers to inspect the reasoning.

\begin{figure}[t]
\centering
\begin{tikzpicture}[font=\scriptsize,>=Latex,n/.style={draw,rounded corners=2pt,align=center,text width=1.8cm,minimum height=.9cm,fill=blue!4}]
\node[n] (impeach) at (0,1.5) {A: DJT impeached\\before 20 Jan 2029};
\node[n] (convict) at (2.9,1.5) {B: DJT impeached and convicted\\before 20 Jan 2029};
\node[n] (resign) at (2.9,-.2) {C: DJT resigns during\\the 2025--2029 term};
\node[n,fill=green!5] (leave) at (5.8,.65) {D: DJT leaves office\\before 20 Jan 2029};
\draw[->,thick] (impeach.east)--node[above,align=center]{+\\medium}(convict.west);
\draw[->,thick] (convict.east)--node[above,sloped]{+, high}(leave.west);
\draw[->,thick] (resign.east)--node[below,sloped]{+, high}(leave.west);
\end{tikzpicture}
\caption{An example semantic relation graph. Arrows indicate the proposed ordering and carry association signs and strength labels. Rationales distinguish the prerequisite relation A to B from the departure mechanisms B to D and C to D. Appendix~\ref{app:example} gives exact descriptions and tickers. Selection of this illustration is independent of correlation performance.}
\Description{Four contract nodes form three positive edges. Impeachment connects to impeachment and conviction with medium strength. Impeachment and conviction, and resignation, each connect to early departure with high strength. Arrow direction represents the ordering.}
\label{fig:example}
\end{figure}
Figure~\ref{fig:example} shows the resulting local graph (excluding other markets in the cluster that would also be related). The removal-to-departure relation, for example, is mapped to
\begin{gather*}
\texttt{KXTRUMPREMOVE} \to \texttt{KXTRUMPOUT27-27-JAN2029}
\end{gather*}
Correlation evaluation uses the validated tickers and association sign; the contract text and rationale explain the semantic basis of each edge.

\section{A Trade-Price Correlation Benchmark}
\label{sec:benchmark}
For assessing behavioral relevance, trade-price histories offer a more direct empirical reference than annotator judgments. They reflect transactions by financially exposed participants and provide repeated observations of how market valuations change as information arrives~\cite{wolfers2004prediction}. Fixed sampling rules and matched controls make the comparison repeatable across graph generators. Human annotation assesses semantic plausibility and rationale quality; price correlation measures how a proposed relation is expressed in observed market behavior.

\subsection{Trade histories and observed prices}
For the monthly catalogs described in Section~\ref{sec:market-sample}, we obtain trade histories containing market tickers, execution timestamps, and Yes prices in dollars~\cite{kalshi2026trades}. Each monthly evaluation uses only trades executed within that UTC calendar month, with prices in $[0,1]$. The resulting input contains \TradeRows{} trade observations across \TradeMarketMonths{} market-months with observed trades.
\emph{Eligible relation pairs} have sufficient overlapping price observations and price variation for correlation analysis. The observation requirements apply jointly to each pair and its control (Section~\ref{sec:control-comparisons}). Table~\ref{tab:coverage} counts monthly relation pairs eligible in at least one configuration at the reported intervals.

\begin{table}[t]
\caption{Monthly market sample and trade-data coverage. Markets, clusters, and edges describe the graph catalogs. Traded counts markets with at least one trade; Trades counts trades in millions. Eligible pairs satisfy the shared-observation and price-variation requirements in at least one configuration at the reported intervals.}
\label{tab:coverage}
\centering\small
\begin{tabular}{lrrrrrr}
\toprule
 & Markets & Clusters & Edges & Traded & Trades & \shortstack{Eligible\\pairs}\\
\midrule

Feb & 6,880 & 41 & 760 & 6,363 & 3.39 & 207 \\
Mar & 6,726 & 53 & 1,740 & 5,623 & 2.28 & 245 \\
Apr & 6,077 & 89 & 370 & 5,565 & 3.22 & 153 \\
May & 6,510 & 108 & 420 & 5,779 & 3.12 & 50 \\
Jun & 8,288 & 116 & 534 & 7,774 & 4.75 & 61 \\
Jul & 7,908 & 97 & 531 & 7,382 & 5.51 & 58 \\
Aug & 9,660 & 91 & 662 & 8,854 & 4.19 & 70 \\
Sep & 13,377 & 90 & 650 & 12,021 & 3.85 & 84 \\

\bottomrule
\end{tabular}
\end{table}

For interval $h\in\{3,12,24\}$ hours, $p_{i,t}^{(h)}$ is the last observed Yes trade price in UTC bin $t$, with empty bins excluded.\footnote{Execution times within a common bin can differ across contracts.}
A price change over $k$ bins is defined as
\begin{equation*}
 d_{i,t}^{(h,k)}=p_{i,t}^{(h)}-p_{i,t-k}^{(h)},\qquad k\in\{1,3\}.
\end{equation*}
Dollar changes respect the bounded probability scale and remain defined at zero.
The main analysis uses one-bin price changes ($k=1$), so the horizon of change equals the sampling interval. Varying this interval from three hours to one day examines co-movement across temporal resolutions~\cite{epps1979}, while holding the change definition fixed. Pearson measures linear association; Spearman evaluates ranked changes using the same controls, timestamp rules, and aggregation.
Three-bin changes provide another longer-horizon robustness check, and may bridge a missing intermediate bin and overlap in time, measuring changes over a longer horizon.

\subsection{Random and matched controls}
\label{sec:control-comparisons}
For edge $(i,j)$, market $i$ is the common anchor; we select up to three control contracts $u$ from its cluster. Control candidates must belong to a different event from $i$, have at least four observed first-week hourly prices, and no known graph link to $i$ in either source. 
Thus, an omitted graph edge represents an unselected pair.
Genuine undiscovered relationships may remain among the controls. The benchmark compares graph-selected pairs with these unselected pairs under the same observation requirements.

We use two baselines: the \textit{random baseline} samples control candidates uniformly, and the \textit{matched baseline} ranks candidates by squared standardized distance from market $j$ on first-week mean price, log observed activity, and log root-mean-square hourly price change. 
To compare the matching features on a common scale, we divide each feature difference by its standard deviation across candidate control markets.
At most 200 control candidates are sampled for each anchor using a stable seed, with the closest three being selected for the matched baseline. 
Appendix~\ref{app:controls} gives the exact feature definitions and sparse-history conventions.

We use the first seven UTC days of each month for \emph{control matching}. Their price, activity, and variability features determine the controls, whose identities are then kept fixed. \emph{LFWO (leave-first-week-out)} measures correlations from day eight through month-end. LFWO separates first-week control matching from subsequent correlation measurement. Full-month evaluation uses the complete calendar month and provides greater observation coverage. Both periods use the same selected controls. Graph construction and specification selection are retrospective.

For each triple $(i,j,u)$, where $i$ is the common anchor market, $j$ is linked to $i$ in the cluster relation graph, and $u$ is a control market assigned to $i$, we calculate $\rho(i,j)$ and $\rho(i,u)$ using \emph{the same timestamp mask}, requiring finite price changes for all three contracts. 
We also require that each window has 12 common observations spanning at least half its length, at least three nonzero changes in each series, and positive variance. 
These requirements determine whether a relation pair has sufficient shared observations and price variation for comparison with that control, and all filtering applies jointly to the triple.

\subsection{Estimation and aggregation}
For relation sign $s_{ij}$, a triple's signed correlation advantage is
\begin{equation*}
 \delta_{iju,w}=s_{ij}\left[\rho_w(i,j)-\rho_w(i,u)\right].
\end{equation*}
We use Pearson correlation for one- and three-bin changes and Spearman correlation for one-bin changes. For a positive edge, the score increases when the linked pair's correlation exceeds the control correlation; for a negative edge, it increases when the linked pair's correlation is more negative. A value of $+0.1$ is an advantage of one tenth of a correlation unit after applying the graph sign to both terms. Separate sign and absolute-correlation sensitivities evaluate polarity assignment and association magnitude.
The evaluation window specifies the period over which each correlation is estimated. \emph{Full month} uses the entire UTC calendar month, including the control-matching week. \emph{LFWO} uses the remaining month to estimate one correlation per pair and control. Both endpoint prices of every included LFWO change must lie on or after the start of day eight.
The \emph{7-day} and \emph{14-day rolling} sensitivities estimate correlations separately in windows of those lengths after the matching week, advancing by seven days and averaging the valid window correlations. 

Windows are averaged within each triple, valid controls within each edge, and edges within each month-cluster. Let $\bar\delta_{mc}$ be that cluster mean and $\mathcal C_m$ the clusters containing eligible relation pairs. The reported effect is
\begin{equation}
 \widehat\Delta=\frac1M\sum_{m=1}^{M}\frac1{|\mathcal C_m|}
       \sum_{c\in\mathcal C_m}\bar\delta_{mc}.
 \label{eq:aggregation}
\end{equation}
Months and their eligible clusters receive equal weight. A market appearing in several months can contribute several edge-month observations; the count describes coverage rather than independent replications. Strength restrictions use the same control assignments and recalculate cluster and month means using eligible pairs in each strength subset.

\subsection{Statistical inference}
\label{sec:uncertainty}

Market pairs share contracts, respond to common news, and recur across months~\cite{cameron2015cluster}. Our main inference uses each month's equal-cluster mean, $\widehat\Delta_m$, as one observation, allowing dependence among all comparisons within that month. Let $s_\Delta$ be the sample standard deviation of the $M$ monthly effects. We report the two-sided month-level statistic
\begin{equation*}
 T=\frac{\widehat\Delta}{s_\Delta/\sqrt M},\qquad
 \widehat\Delta\ \pm\ t_{M-1,0.95}\frac{s_\Delta}{\sqrt M}.
\end{equation*}
The second expression gives the pointwise 90\% confidence interval. The main table uses a Student $t$ reference with $M-1$ degrees of freedom; its comparisons have $M=8$ and seven degrees of freedom. This grouped approach assumes approximately independent, Gaussian monthly estimates~\cite{ibragimov2010}; recurring markets and serial news transmission are relevant to that assumption.

We apply the same test and interval to Pearson and Spearman, all three intervals, both measurement periods, and each strength restriction. We conduct a retrospective specification analysis across these choices~\cite{simonsohn2020}. The $p$-values describe exploratory evidence for individual settings. Appendix~\ref{app:inference} gives the evaluation scope and complementary month-level checks.

\section{Results}
\label{sec:results}
\begin{figure*}[t]
\centering\includegraphics[width=\textwidth]{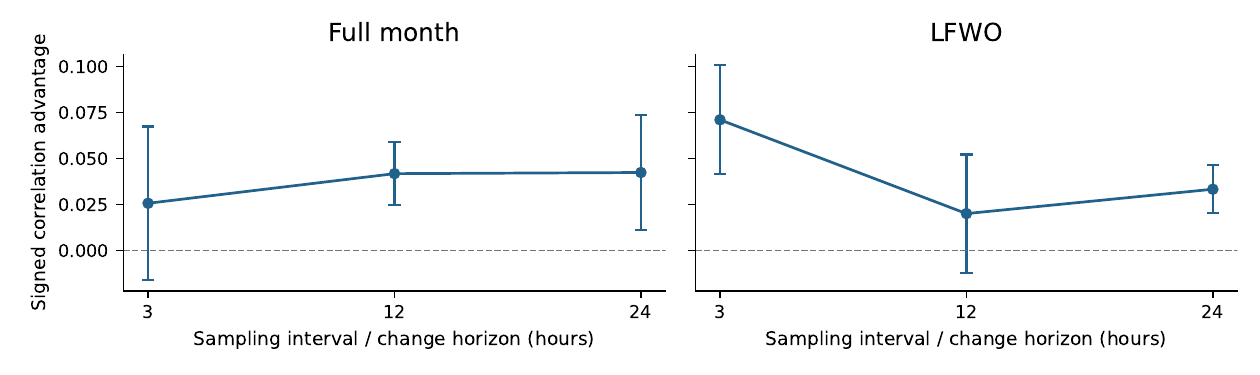}
\caption{Signed Pearson price-change correlation advantages for all strengths at 3-, 12-, and 24-hour intervals against matched controls. Full month uses the complete calendar month; LFWO uses the period after the first control-matching week. Bars are pointwise 90\% month-level $t$-intervals corresponding to the tests in Table~\ref{tab:systematic-configurations}.}
\Description{Two panels show all-strength signed Pearson correlation advantages at 3, 12, and 24 hours for Full month and LFWO evaluation. Every point has a 90 percent month-level t confidence interval.}
\label{fig:frequencies}
\end{figure*}

The primary results report signed Pearson price-change correlation advantages for all strengths at 3, 12, and 24 hours under Full month and LFWO evaluation. Table~\ref{tab:systematic-configurations} gives the effects and two-sided month-level $p$-values; Figure~\ref{fig:frequencies} shows the corresponding pointwise 90\% intervals. A three-hour LFWO comparison examines rank association and relation strength under the same controls and observation rules. Appendix~\ref{app:robustness} provides the complete Spearman and strength analyses.

\begin{table}[t]
\caption{Signed Pearson price-change correlation advantages, all strengths, against matched controls. Hours gives sampling interval and change horizon; $\Delta$ is signed correlation advantage; $N$ counts eligible edge-months. All $p$-values are two-sided month-level $t$-test values with seven degrees of freedom. \significancenote}
\label{tab:systematic-configurations}
\centering\small
\begin{tabular}{llrrr}
\toprule
Period & Hours & $\Delta$ & $p$ & $N$\\
\midrule

Full month & 3 & +0.026 & $0.2813$ & 342 \\
Full month & 12 & +0.042 & $0.0024^{***}$ & 599 \\
Full month & 24 & +0.042 & $0.0369^{**}$ & 692 \\
\addlinespace
LFWO & 3 & +0.071 & $0.0027^{***}$ & 301 \\
LFWO & 12 & +0.020 & $0.2739$ & 517 \\
LFWO & 24 & +0.033 & $0.0019^{***}$ & 595 \\

\bottomrule
\end{tabular}
\end{table}

\subsection{Temporal resolution}
Table~\ref{tab:systematic-configurations} shows positive full-month Pearson advantages at each reported interval, ranging from $+0.026$ to $+0.042$. The 12-hour effect is positive in all eight months ($p=0.0024$); the daily effect is positive in seven ($p=0.0369$). These comparisons link the graph's cross-event connections to price associations beyond matched within-topic controls. Each interval measures ordinary one-bin changes, giving the same interpretation to its sampling interval and change horizon.

Under LFWO, three-hour Pearson gives $+0.071$ on 301 edge-months across 97 month-clusters, with seven positive months ($p=0.0027$). Daily changes give $+0.033$ on 595 edge-months, positive in all eight months ($p=0.0019$). These settings demonstrate complementary strengths: a larger three-hour association advantage and consistent daily effects across the sample period. Figure~\ref{fig:frequencies} places these findings within the complete interval sequence and shows the variation in precision across settings.

\subsection{Robustness checks}
\label{sec:strength}
Table~\ref{tab:robustness} keeps the three-hour LFWO design fixed and changes the association statistic and strength restriction. Spearman yields an advantage of $+0.073$ ($p=0.0131$), supporting the finding through ranked price changes. Medium/high relations yield $+0.107$ under Pearson ($p=0.0112$) and $+0.100$ under Spearman ($p=0.0282$). All four comparisons are positive in seven of eight months. The restriction includes 223 of the 301 eligible edge-months and increases the observed advantage by $0.036$ under Pearson and $0.028$ under Spearman.

Medium/high strength labels identify pairs with larger correlation advantages under both Pearson and Spearman. Appendix~\ref{app:robustness} reports the full interval and strength grids, and Appendix~\ref{app:grid} fixes eligible edge membership across intervals. On that common cohort, three-hour LFWO Spearman gives $+0.072$.

\begin{table}[t]
\caption{Three-hour LFWO robustness against matched controls. $\Delta$ is signed correlation advantage; $N$ counts eligible edge-months. All $p$-values are two-sided month-level $t$-test values with seven degrees of freedom. \significancenote}
\label{tab:robustness}
\centering\small\setlength{\tabcolsep}{4pt}
\begin{tabular}{llrrr}
\toprule
Statistic & Strength & $\Delta$ & $p$ & $N$\\
\midrule
Pearson & All & +0.071 & $0.0027^{***}$ & 301 \\
Pearson & Medium/high & +0.107 & $0.0112^{**}$ & 223 \\
Spearman & All & +0.073 & $0.0131^{**}$ & 301 \\
Spearman & Medium/high & +0.100 & $0.0282^{**}$ & 223 \\
\bottomrule
\end{tabular}
\end{table}

\section{Discussion}
\label{sec:discussion}
Our piepeline organizes contract descriptions into topic clusters, proposes cross-event relations through structured reasoning, and anchors each assertion to an exact tradable proposition. The behavioral benchmark evaluates whether a relation graph captures signed price associations beyond matched within-topic controls. For the evaluated graph, three-hour Pearson and Spearman yield positive signed correlation advantages, with larger point estimates for medium/high relations. These findings provide behavioral evidence for graph-selected links under the stated controls and observation rules. Controlled comparisons between graph generators can quantify the contribution of individual construction steps.

The worked example shows how this structure becomes interpretable. A resolution rule fixes the proposition, a ticker fixes its identity, and a rationale explains the proposed mechanism. The removal and resignation edges express distinct routes to early departure; the impeachment edge indicates a prerequisite with an additional condition. 
Each proposed relation can be checked against the contracts' resolution rules.

The resulting benchmark supports controlled comparisons of graph-construction methods. Applying the same model and schema to direct pairs on a fixed contract universe can quantify the contribution of grouping and batch reasoning. Global candidate retrieval offers a complementary comparison, and blinded annotation can measure semantic precision and rationale faithfulness.

Explicit contract identities, relation signs, and rationales support inspection of the semantic assertions alongside behavioral evaluation. The generating LLM's documented pre-February-2025 knowledge horizon and exclusion of price inputs define its information boundary. LFWO separately partitions control matching and correlation measurement. A prospective study would additionally fix graph membership, metadata, and evaluation specifications before observing the measurement period. Section~\ref{sec:uncertainty} specifies the retrospective inference scope used here.

\section{Conclusion}
Our pipeline turns contract language into a structured graph of cross-event relationships with exact identities, resolution conditions, and inspectable rationales. The accompanying benchmark tests the behavioral relevance of graph-selected links through signed price-correlation advantages over within-topic controls. For the evaluated graph, three-hour Pearson advantages under LFWO are $+0.071$ overall and $+0.107$ for medium/high relations; Spearman gives $+0.073$ and $+0.100$. All four comparisons are positive in seven of eight months. Full-month 12-hour Pearson advantages and daily Pearson advantages under LFWO are positive in all eight months. 
Overall, our pipeline and the benchmark make the relationships implicit in prediction-market contracts explicit, inspectable, and empirically testable against the collective beliefs expressed in market prices.


\label{end-main-text}
\bibliographystyle{ACM-Reference-Format}
\bibliography{references}
\clearpage
\appendix
\input{appendix}
\end{document}

%% file: appendix.tex
\section{Graph Construction Details}
\label{app:provenance}
We construct the graph as follows. Its batch boundary defines which contracts the model considers jointly; union and validation assemble those decisions into the graph.
\begin{quote}\small
\textbf{Input:} monthly contract catalog $X$; batch size $b=10$; model $\theta$; typed edge schema.\\
1. Embed unique event titles and assign topic clusters.\\
2. For each cluster $C$, partition its ordered contracts into batches $B_k$ of size at most $b$.\\
3. Apply the typed LLM transducer to each $B_k$ with event title, question, and primary resolution rule as inputs. Allow an empty relation list.\\
4. For every proposed edge, resolve its endpoint texts to unique tickers and enforce the event and cluster eligibility rules.\\
5. Canonicalize each unordered pair, collapse duplicates, and exclude sign conflicts.\\
\textbf{Output:} signed relation edges with strength and rationale, indexed by month and source.
\end{quote}

The generation prompt asks the model to select related pairs, reproduce their endpoint descriptions exactly, assign a Boolean for likely same/different outcomes, provide a low/medium/high strength, and justify each choice. It requires different event titles and imposes no edge quota. This schema supports a selective graph while distinguishing an outcome judgment from a measured price correlation.

\subsection{Market selection query}
\label{app:sample}
The market-selection query selects reports whose timestamps fall between \texttt{DateStart} and \texttt{DateStop}, with both endpoints included. It groups by ticker and category, sums \texttt{daily\_volume}, and selects groups with total volume at least 10,000. The status filter uses the group's maximum reported status string, as implemented by \texttt{MAX(status)}; separate calls request \texttt{active} and \texttt{finalized}. The query excludes the report categories Crypto, Sports, Exotics, Unknown, and Financials, and tickers containing \texttt{KXHIGH}, \texttt{KXLOWT}, or \texttt{KXRAIN}. Results are ordered by total volume and limited to 10,000 ticker--category groups per call. The date predicate concerns report timestamps; trade-price evaluation uses the UTC calendar-month restriction in Section~\ref{sec:benchmark}.

The category and volume conditions define the initial market sample. Event enrichment retrieves sibling contracts and API metadata, so individual contracts in the expanded catalogs can have lower reported volume, and API category labels can differ from the source-report categories. For example, the April report classifies ticker \texttt{KXROCKANDROLLHALLOFFAME-}\allowbreak\texttt{26-MAR} as Entertainment; API enrichment labels its event Sports. Selection uses the report table's taxonomy. The monthly counts describe the clustered catalogs assembled retrospectively after event enrichment.

\subsection{Worked-example details}
\label{app:example}
Table~\ref{tab:example-inputs} gives the four input descriptions behind Figure~\ref{fig:example}. The three illustrated edges are LLM outputs for the March 2026 catalog: B to D and C to D have \texttt{direction=true} and high strength; A to B has \texttt{direction=true} and medium strength. The node letters are presentation aids. Contract identity is fixed by the ticker, and a description match uses the complete question and rule.

\begin{table*}[t]
\centering\small
\caption{Exact contract descriptions for the worked construction example. Questions and primary resolution rules are reproduced from the relation endpoints. Each row maps to a unique exchange ticker.}
\label{tab:example-inputs}
\begin{tabular}{@{}p{.035\textwidth}p{.235\textwidth}p{.69\textwidth}@{}}
\toprule
Node & Ticker & Question and primary resolution rule\\
\midrule
A & \texttt{KXIMPEACH-29-JAN20} & Will the President be impeached before Jan 20, 2029? If the President of the United States has been impeached before Jan 20, 2029, then the market resolves to Yes.\\[4pt]
B & \texttt{KXTRUMPREMOVE} & Will Trump be impeached and removed from office? If the President of the United States has been impeached and convicted by the U.S. Senate before Jan 20, 2029, then the market resolves to Yes.\\[4pt]
C & \texttt{KXTRUMPRESIGN} & Will President Trump resign before his term is up? If the President of the United States elected for the 2025-2029 term resigns the office, then the market resolves to Yes.\\[4pt]
D & \texttt{KXTRUMPOUT27-27-}\newline\texttt{JAN2029} & Will Donald Trump leave office before January 20, 2029? If Donald Trump leaves office before January 20, 2029, then the market resolves to Yes.\\
\bottomrule
\end{tabular}
\end{table*}

For B to D, \texttt{market\_info\_leader} contains B's complete description. The field \texttt{market\_info\_follower} contains D's. The Boolean \texttt{direction} is true, and \texttt{strength} is high. The rationale reads:
\begin{quote}\small
If Trump is impeached and convicted by the Senate, he will be removed from office, meaning he will leave office before January 20, 2029. The leader market resolving to Yes guarantees the follower market resolves to Yes.
\end{quote}
The rationale asserts an implication from removal to departure. The sign specifies the proposed association; conditional probability and temporal price leadership require separate estimation. For A to B, the rationale identifies impeachment as a necessary prerequisite for removal and predicts a positive association. These rationales distinguish two mechanisms behind edges with the same sign.

\section{Control Features and Observation Rules}
\label{app:controls}
Let $p_{i,t}^{(1)}$ be an observed hourly price in the first seven UTC days. The features for market $i$ are its observed mean price, $\log(1+n_i)$ with $n_i$ observed first-week hourly prices, and $\log(v_i+10^{-5})$. Here
\begin{equation*}
 v_i=\sqrt{\frac{\sum_{t\in\mathcal O_i}(p_{i,t}^{(1)}-p_{i,t-1}^{(1)})^2}
                         {\max(1,|\mathcal O_i|)}},
\end{equation*}
and $\mathcal O_i$ requires both adjacent prices. Means use a denominator of at least one, so a history with no observations has mean and RMS change zero. Such a market can occur as a relation endpoint; control candidates require $n_i\geq4$. The convention permits matching for a sparse endpoint while limiting the interpretation of feature balance in that case.

Squared feature distance divides each coordinate by its standard deviation among control candidates. Coordinates with scale below $10^{-8}$ use scale one. The candidate cap is 200, and the maximum number of controls per edge is three. Stable hash-derived seeds make control sampling reproducible; the configuration seed is 20261022. Matches are selected at hourly frequency once and held fixed across all measurement specifications.

Eligible relation pairs have sufficient overlapping price observations and price variation for correlation analysis with a control. For each triple, a missing observation in any of its three series removes that timestamp from both correlations. At least 12 shared observations must span at least half the window. Each change series must contain three changes exceeding $10^{-10}$ in absolute value. The correlation denominator must exceed $10^{-12}$. Spearman ranks are computed within the common mask. Per-edge relation scores can vary across controls because their common masks vary. Scores are averaged using the sequence in Equation~\eqref{eq:aggregation}.

\section{Correlation and Strength Robustness}
\label{app:robustness}
\label{app:strength-grid}
\begin{figure*}[t]
\centering\includegraphics[width=\textwidth]{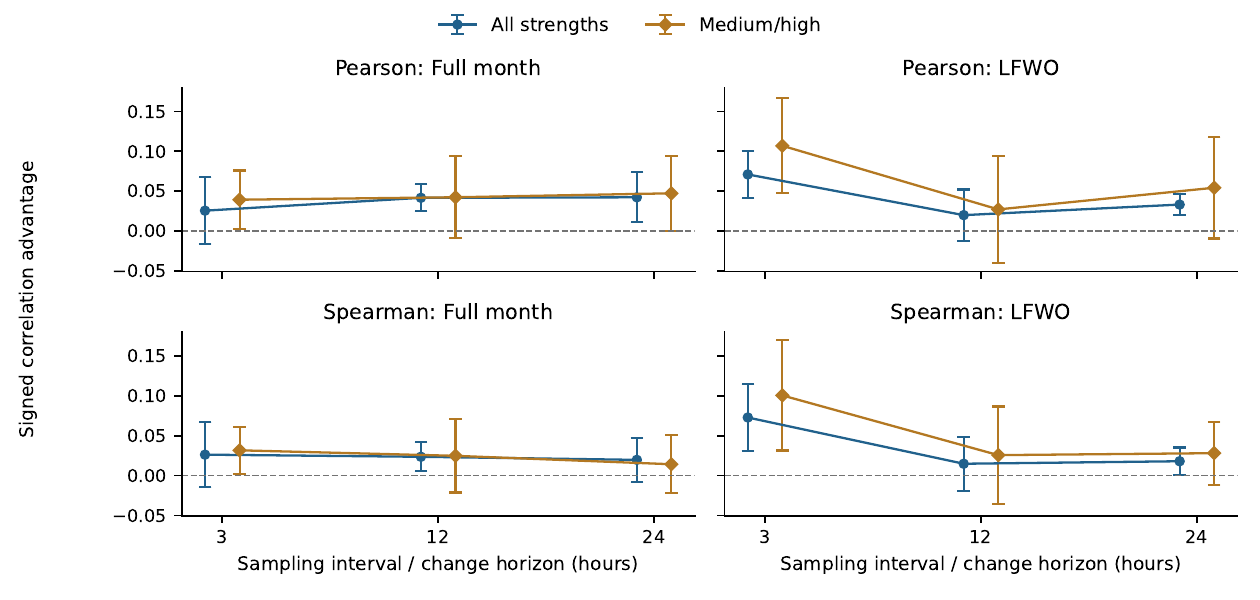}
\caption{Pearson and Spearman robustness against matched controls, for all strengths and medium/high relations. Columns use Full month and LFWO; rows use Pearson and Spearman. Bars show pointwise 90\% month-level $t$-intervals. Table~\ref{tab:full-correlations} gives the corresponding $p$-values and high-only results.}
\Description{Four panels compare Pearson and Spearman signed correlation advantages at 3, 12, and 24 hours for all strengths and medium/high relations. The two evaluation periods are Full month and LFWO, with 90 percent month-level t intervals.}
\label{fig:full-robustness}
\end{figure*}

Table~\ref{tab:full-correlations} and Figure~\ref{fig:full-robustness} evaluate Pearson and Spearman at 3-, 12-, and 24-hour intervals under Full month and LFWO. All-strength and medium/high comparisons cover all eight months. The high-only three-hour settings cover seven months; the high-only 12- and 24-hour settings cover eight. Each restriction uses the same control assignments and recomputes cluster and month means over the selected edges. This applies the same benchmark to each strength subset, allowing comparison of effect size and precision.

\begin{table*}[t]
\caption{Complete correlation and strength results against matched controls. Hours gives sampling interval and change horizon; $\Delta$ is signed correlation advantage; $N$ counts eligible edge-months. All $p$-values are two-sided month-level $t$-test values. All and medium/high use eight months; high-only uses seven months at three hours and eight at the other intervals. \significancenote}
\label{tab:full-correlations}
\label{tab:high-grid}
\centering\small\setlength{\tabcolsep}{4pt}
\begin{tabular}{llrrrrrrrrr}
\toprule
 & & \multicolumn{3}{c}{All strengths} & \multicolumn{3}{c}{Medium/high} & \multicolumn{3}{c}{High only}\\
Period & Hours & $\Delta$ & $p$ & $N$ & $\Delta$ & $p$ & $N$ & $\Delta$ & $p$ & $N$\\
\midrule
\multicolumn{11}{l}{\textit{Pearson price-change correlation}} \\
Full month & 3 & +0.026 & $0.2813$ & 342 & +0.039 & $0.0804^{*}$ & 244 & +0.137 & $0.0852^{*}$ & 130 \\
Full month & 12 & +0.042 & $0.0024^{***}$ & 599 & +0.042 & $0.1596$ & 373 & +0.029 & $0.5083$ & 192 \\
Full month & 24 & +0.042 & $0.0369^{**}$ & 692 & +0.047 & $0.0985^{*}$ & 410 & +0.058 & $0.4058$ & 215 \\
\addlinespace
LFWO & 3 & +0.071 & $0.0027^{***}$ & 301 & +0.107 & $0.0112^{**}$ & 223 & +0.213 & $0.0103^{**}$ & 120 \\
LFWO & 12 & +0.020 & $0.2739$ & 517 & +0.027 & $0.4665$ & 342 & +0.051 & $0.4387$ & 177 \\
LFWO & 24 & +0.033 & $0.0019^{***}$ & 595 & +0.054 & $0.1486$ & 368 & +0.133 & $0.2699$ & 194 \\
\addlinespace
\multicolumn{11}{l}{\textit{Spearman price-change correlation}} \\
Full month & 3 & +0.026 & $0.2590$ & 342 & +0.032 & $0.0819^{*}$ & 244 & +0.129 & $0.0975^{*}$ & 130 \\
Full month & 12 & +0.024 & $0.0442^{**}$ & 599 & +0.025 & $0.3399$ & 373 & +0.006 & $0.8758$ & 192 \\
Full month & 24 & +0.020 & $0.2216$ & 692 & +0.014 & $0.4798$ & 410 & +0.040 & $0.5363$ & 215 \\
\addlinespace
LFWO & 3 & +0.073 & $0.0131^{**}$ & 301 & +0.100 & $0.0282^{**}$ & 223 & +0.187 & $0.0226^{**}$ & 120 \\
LFWO & 12 & +0.015 & $0.4304$ & 517 & +0.026 & $0.4503$ & 342 & +0.025 & $0.7133$ & 177 \\
LFWO & 24 & +0.018 & $0.0873^{*}$ & 595 & +0.028 & $0.2184$ & 368 & +0.088 & $0.4174$ & 194 \\
\bottomrule
\end{tabular}
\end{table*}

Under three-hour LFWO, Pearson gives $+0.107$ for medium/high relations and $+0.071$ overall. Spearman gives $+0.100$ and $+0.073$. Full-month Pearson means are also larger in the restricted subset at each interval. These comparisons connect the graph's strength attribute to observed associations and document the resulting coverage.

\section{Signs, Baselines, and Monthly Stability}
\label{app:sign}
Table~\ref{tab:monthly} gives monthly effects for the three-hour LFWO robustness comparison. Each cell is an equal-cluster mean within the stated month. Seven of eight monthly effects are positive for both statistics and both strength subsets. The reported overall estimates average these monthly means equally.

\begin{table*}[t]
\caption{Monthly three-hour LFWO effects against matched controls. MH denotes medium/high strength. $N$ counts eligible edge-months, shared by the two statistics within each strength subset.}
\label{tab:monthly}
\centering\small
\begin{tabular}{lrrrrrr}
\toprule
 & \multicolumn{2}{c}{Pearson advantage} & \multicolumn{2}{c}{Spearman advantage} & \multicolumn{2}{c}{Edge-months}\\
Month & All & MH & All & MH & $N_{\mathrm{All}}$ & $N_{\mathrm{MH}}$\\
\midrule
Feb & +0.038 & +0.065 & +0.031 & +0.038 & 90 & 67 \\
Mar & +0.133 & +0.120 & +0.157 & +0.132 & 74 & 66 \\
Apr & +0.085 & +0.076 & +0.080 & +0.074 & 54 & 51 \\
May & +0.042 & +0.074 & +0.012 & +0.026 & 11 & 9 \\
Jun & +0.109 & +0.192 & +0.114 & +0.111 & 14 & 8 \\
Jul & -0.005 & -0.034 & -0.029 & -0.030 & 23 & 9 \\
Aug & +0.093 & +0.100 & +0.116 & +0.142 & 16 & 8 \\
Sep & +0.073 & +0.262 & +0.101 & +0.311 & 19 & 5 \\
\bottomrule
\end{tabular}
\end{table*}

The all-positive comparison preserves endpoints, windows, and observation masks while assigning every edge sign $+1$. The absolute score takes the absolute value of each window/triple correlation before aggregation. Table~\ref{tab:sign} applies both comparisons to three-hour LFWO. They distinguish alignment with the graph's proposed sign from the magnitude of association among its selected endpoints.

\begin{table}[ht]
\caption{Three-hour LFWO polarity and magnitude on the same eligible comparisons. All $p$-values are two-sided month-level $t$-test values. \significancenote}
\label{tab:sign}
\centering\small\setlength{\tabcolsep}{3pt}
\begin{tabular}{llrrrr}
\toprule Statistic & Score & Relation & Control & $\Delta$ & $p$\\
\midrule
Pearson & Graph signs & 0.045 & -0.026 & +0.071 & $0.0027^{***}$ \\
Pearson & All positive & 0.043 & -0.016 & +0.059 & $0.0193^{**}$ \\
Pearson & Absolute & 0.193 & 0.185 & +0.009 & $0.5885$ \\
Spearman & Graph signs & 0.042 & -0.031 & +0.073 & $0.0131^{**}$ \\
Spearman & All positive & 0.052 & -0.026 & +0.078 & $0.0114^{**}$ \\
Spearman & Absolute & 0.182 & 0.177 & +0.005 & $0.7194$ \\
\bottomrule
\end{tabular}
\end{table}

\section{Change Horizons and Evaluation Windows}
\label{app:grid}
\begin{figure*}[t]
\centering\includegraphics[width=.95\textwidth]{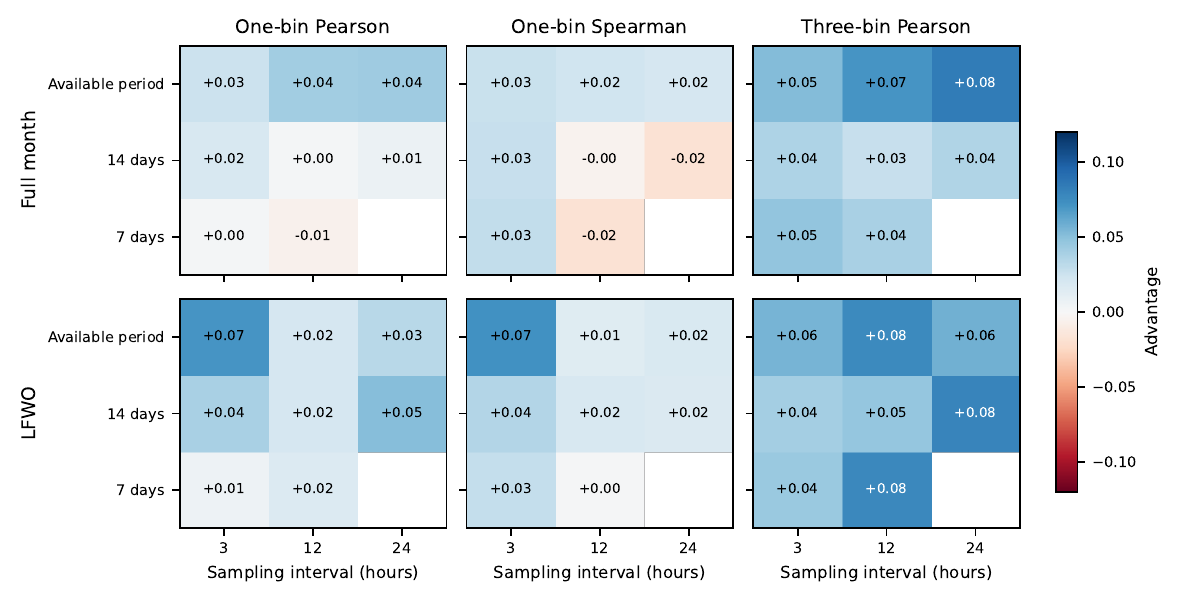}
\caption{Price-change sensitivities against matched controls at 3-, 12-, and 24-hour sampling intervals. Available period uses the complete calendar month in the Full month panels and the remaining month in the LFWO panels. The 7- and 14-day windows advance weekly. The color scale is shared across all panels.}
\Description{Six heatmaps show signed correlation advantages at 3, 12, and 24 hours for three window sizes and three price-change metrics under Full month and LFWO.}
\label{fig:grid}
\end{figure*}

Figure~\ref{fig:grid} evaluates one-bin Pearson, one-bin Spearman, and three-bin Pearson changes at 3-, 12-, and 24-hour sampling intervals. Each specification uses its own eligible relation pairs. Available period means the full calendar month for Full month and the remaining month for LFWO. The 7- and 14-day rows use windows of those lengths, advancing by seven days and averaging valid window correlations. Three-bin changes extend the change horizon to 9, 36, and 72 hours, respectively.

A common-cohort check uses pairs eligible at all three intervals within each measurement regime and metric. With 289 edge-months, full-month three-bin Pearson advantages at 3, 12, and 24 hours are $0.057$, $0.090$, and $0.111$. With 298 edge-months, LFWO one-bin Spearman advantages are $0.072$, $0.019$, and $0.019$. The three-hour rank-association advantage persists with edge membership fixed. Masks and change horizons follow their interval-specific definitions.

Removing any one month leaves the three-hour LFWO Pearson advantage between $+0.062$ and $+0.082$, and the Spearman advantage between $+0.061$ and $+0.087$. Both pooled advantages remain positive under every single-month omission. Together, the monthly table, common-cohort comparison, and omission checks characterize the temporal stability of the graph's behavioral associations.

\section{Month-Level Inference}
\label{app:inference}
The primary test uses the dispersion of the eight monthly effects and a Student $t$ reference with seven degrees of freedom. Exact month sign flips provide a complementary check: all effects in a month change sign together, preserving dependence within that month. Under independent, symmetric monthly effects, all $2^8=256$ sign patterns form the null reference. Two-sided values are $0.0156$ for three-hour LFWO Pearson, $0.0078$ for full-month 12-hour Pearson, $0.0078$ for daily LFWO Pearson, and $0.0234$ for three-hour LFWO Spearman.

A sensitivity to dependence between adjacent months averages February--March, April--May, June--July, and August--September, using a $t$ reference with three degrees of freedom. Two-sided values are $0.0029$ for three-hour LFWO Pearson, $0.0437$ for full-month 12-hour Pearson, and $0.0356$ for daily LFWO Pearson. This sensitivity assumes approximately independent, Gaussian block means. Exact four-block sign flips have a minimum two-sided value of $0.125$.

The retrospective evaluation comprises 1,344 combinations of source, statistic, period, controls, and score. Across this search, every eight-month setting has a Holm-adjusted value above $0.05$~\cite{holm1979}. Displayed $p$-values characterize individual settings. The reported settings and inference methods are data-informed, and the comparisons are exploratory.